\documentclass[10pt,twocolumn,letterpaper]{article}

\usepackage{iccv}
\usepackage{times}
\usepackage{epsfig}
\usepackage{graphicx}
\usepackage{amssymb}
\usepackage{multicol}
\usepackage{multirow}
\usepackage{amsmath}
\usepackage{mathtools}
\usepackage{subfigure}
\usepackage{tabularx}
\usepackage{tabulary}
\usepackage{authblk}
\newcolumntype{x}[1]{>{\centering\arraybackslash}p{#1pt}}

\usepackage{color}

\renewcommand{\ll}{\mathcal{L}}

\definecolor{citecolor}{RGB}{34,196,34}  
\usepackage[pagebackref=true,breaklinks=true,letterpaper=true,colorlinks,citecolor=citecolor,bookmarks=false]{hyperref}

\newlength\savewidth\newcommand\shline{\noalign{\global\savewidth\arrayrulewidth
  \global\arrayrulewidth 1pt}\hline\noalign{\global\arrayrulewidth\savewidth}}
\newcommand{\tablestyle}[2]{\setlength{\tabcolsep}{#1}\renewcommand{\arraystretch}{#2}\centering\footnotesize}

\newcommand{\tabincell}[2]{\begin{tabular}{@{}#1@{}}#2\end{tabular}}

\renewcommand{\paragraph}[1]{\noindent\textbf{#1}}

\iccvfinalcopy 

\def\iccvPaperID{5695} 

\ificcvfinal\fi

\begin{document}

\title{ SO-Pose: Exploiting Self-Occlusion for Direct 6D Pose Estimation}

\author[1]{Yan Di \thanks{Codes will be released at https://github.com/shangbuhuan13/SO-Pose}}
\author[2]{Fabian Manhardt}
\author[3]{Gu Wang}
\author[3]{Xiangyang Ji}
\author[1]{Nassir Navab}
\author[1,2]{Federico Tombari}
\affil[1]{Technical University of Munich, \textsuperscript{2}Google, \textsuperscript{3}Tsinghua University}
\affil[*]{\tt\small shangbuhuan13@gmail.com, fabianmanhardt@google.com, nassir.navab@tum.de, wangg16@mails.tsinghua.edu.cn, xyji@tsinghua.edu.cn,
tombari@in.tum.de}

\renewcommand\Authands{ and }

\maketitle
\ificcvfinal\thispagestyle{empty}\fi

\begin{abstract}
Directly regressing all 6 degrees-of-freedom (6DoF) for the object pose (\ie the 3D rotation and translation) in a cluttered environment from a single RGB image is a challenging problem. 
While end-to-end methods have recently demonstrated promising results at high efficiency, they are still inferior when compared with elaborate P$n$P/RANSAC-based approaches in terms of pose accuracy. 
In this work, we address this shortcoming by means of a novel reasoning about self-occlusion, in order to establish a two-layer representation for 3D objects which considerably enhances the accuracy of end-to-end 6D pose estimation. 
Our framework, named SO-Pose, takes a single RGB image as input and respectively generates 2D-3D correspondences as well as self-occlusion information harnessing a shared encoder and two separate decoders. 
Both outputs are then fused to directly regress the 6DoF pose parameters. 
Incorporating cross-layer consistencies that align correspondences, self-occlusion and 6D pose, we can further improve accuracy and robustness, surpassing or rivaling all other state-of-the-art approaches on various challenging datasets.
\end{abstract}

\section{Introduction}
Estimating the 6D pose has been widely adopted as an essential cue in high-level computer vision tasks, including robotic grasping and planning~\cite{azad2007stereo}, augmented reality~\cite{tan2018real}, and autonomous driving~\cite{manhardt2019roi,qi2018frustum}. 
Driven by the recent success of deep learning, current methods are capable of estimating the 6D pose in a cluttered environment at impressive accuracy and high efficiency~\cite{Kehl2017,zakharov2019dpod,labbe2020cosypose}. 
Almost all current top-performing frameworks adopt a two-stage strategy that resorts to first establishing 2D-3D correspondences and then computing the 6D pose with a RANSAC-based Perspective-n-Point (P$n$P) algorithm~\cite{zakharov2019dpod,li2019cdpn,hodan2020epos}. Nevertheless, while achieving great results, these methods cannot be trained in an end-to-end manner and require extra computation for optimization of pose. Moreover, adopting surrogate training losses instead of directly predicting 6D poses also prevents further differentiable processing/learning (\eg~by means of self-supervised learning~\cite{wang2020self6d}) and does not allow to incorporate other down-stream tasks. 

\begin{figure}[t!]
\centering
\subfigure[Basic structure of baseline methods \cite{hu2020single, GDRN}]{
\label{Fig.sub.1}
\includegraphics[width=0.99\linewidth]{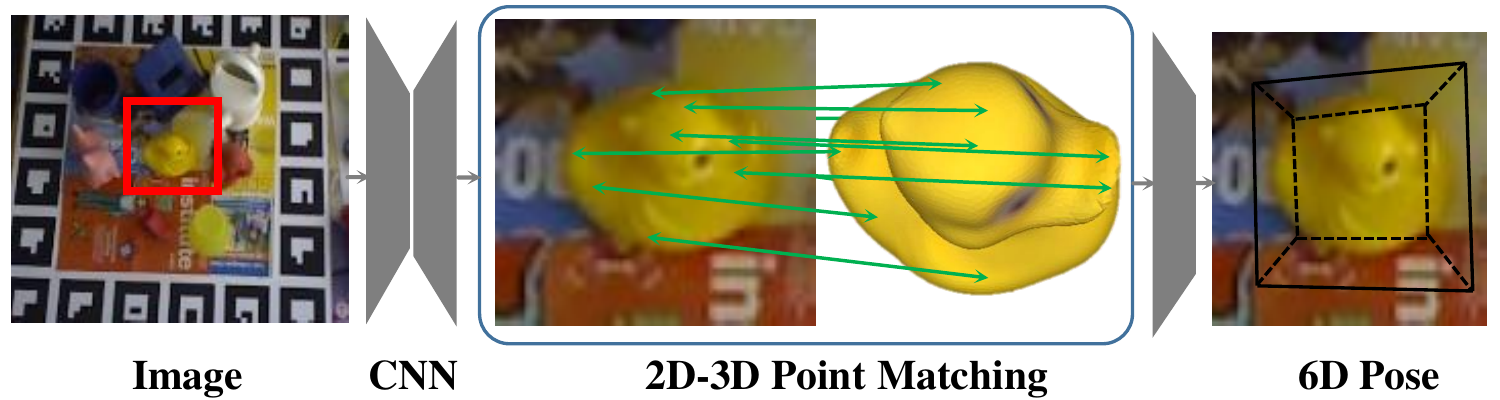}}
\subfigure[Basic structure of our method SO-Pose.]{
\label{Fig.sub.2}
\includegraphics[width=0.99\linewidth]{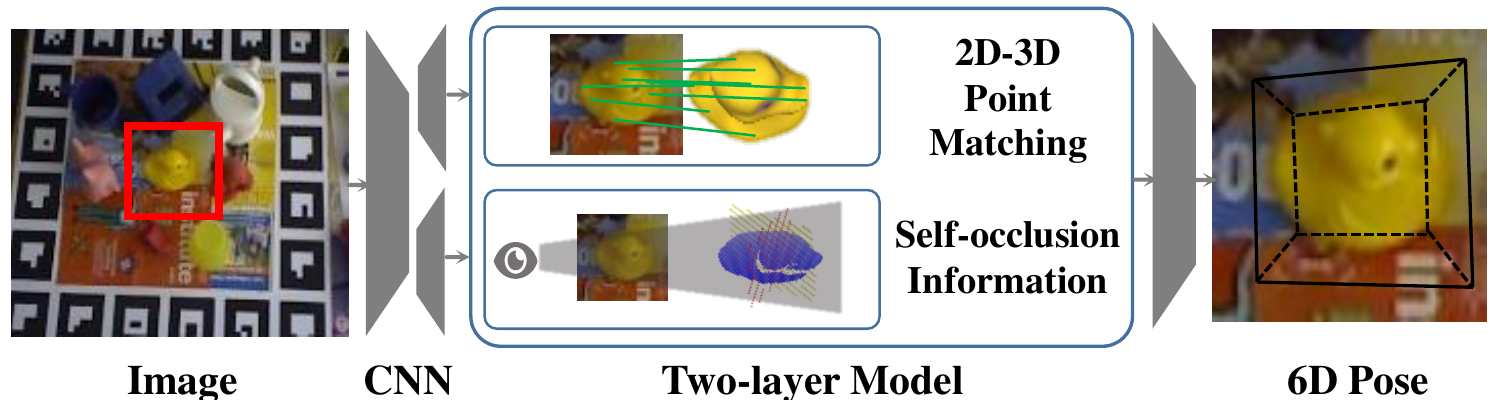}}
\caption{\textbf{The basic structures of end-to-end 6D pose estimation methods.}
Compared to single-layer methods \cite{hu2020single, GDRN} that depend on 2D-3D point matching as intermediate results, our method SO-Pose presents a novel two-layer representation that additionally incorporates self-occlusion information about the object.}
\label{Fig.teaser}
\end{figure}

Despite two-stage approaches dominating the field, a few methods conducting end-to-end 6D pose estimation have been also recently proposed~\cite{hu2020single,chen2020end,wang2020self6d, GDRN}.
They typically learn the 6D pose directly from dense correspondence-based intermediate geometric representations, as shown in Fig.~\ref{Fig.sub.1}.
Nevertheless, although end-to-end methods keep constantly improving, they are still far inferior to two-stage methods harnessing  multi-view consistency check~\cite{labbe2020cosypose}, symmetry analysis~\cite{hodan2020epos}, or disentangled predictions~\cite{li2019cdpn}. 

What limits the accuracy of end-to-end methods? 
After in-depth investigation in challenging scenes, we observe that while the network is approaching the optimum, due to the inherent matching ambiguity of textureless object surface, mis-matching error caused by noise is inevitable, resulting often in one correspondence field corresponding to many 6D poses with similar fitting errors.
This leads the training process to converge to a sub-optimum, hindering the overall 6D pose estimation performance.
Since eliminating errors caused by noise is not trivial, an alternative solution to this problem is to replace the correspondence field with a more precise representation of the 3D object, thus reducing the influence of noise. 

In this work, we attempt at closing the gap between end-to-end and two-stage approaches by leveraging self-occlusion information about the object.
For an object in 3D space, we can logically only observe its visible parts due to the nature of the perspective projection. 
Yet, parts that are invisible due to (self-) occlusion are usually neglected during inference. 
Inspired by multi-layer models used in 3D reconstruction~\cite{shin2019multilayer}, we focus on self-occlusion information to establish a viewer-centered two-layer representation of the object pose.
While the first layer preserves the correspondence field of visible points on the object and their projections, the second layer incorporates the self-occlusion information. 
In essence, instead of directly identifying whether and where each visible point occludes the object, we simplify the procedure by examining the self-occlusion between each pixel and the object coordinate planes.
As illustrated in Fig.~\ref{fig_pro_selfocc}, the ray passing through the camera center and each visible point intersects the object coordinate plane at most three different locations.
The coordinates of these intersections are then utilized to form the second layer representation of the object, as shown in Fig.~\ref{Fig.sub.2}.
Finally, two cross-layer consistency losses are introduced to align self-occlusion, correspondence field and 6D pose simultaneously, reducing the influence of noise.

To summarize, our main contributions are as follows:
\begin{itemize}
\setlength{\itemsep}{0pt}
\setlength{\parsep}{0pt}
\setlength{\parskip}{0pt}
\item We propose SO-Pose, a novel deep architecture that directly regresses the 6D pose from the two-layer representation of each 3D object.
\item We propose to leverage self-occlusion and 2D-3D correspondences to establish a two-layer representation for each object in 3D space, which can be utilized to enforce two cross-layer consistencies.
\item SO-Pose consistently surpasses all other end-to-end competitors on various challenging datasets. Moreover, SO-Pose also achieves comparable accuracy when compared with other state-of-the-art two-stage methods, whilst being much faster.
\end{itemize}
\section{Related Works}
The related works for monocular 6D pose estimation can be roughly partitioned into three different lines of works. 
In particular, while some methods directly regress the final 6D pose, others either learn a latent embedding for subsequent retrieval of the pose, or employ 2D-3D correspondences to solve for the 6D pose by means of the well-established RANSAC/P$n$P paradigm.

As for the first line of works, Kehl \etal~\cite{Kehl2017} extends SSD~\cite{liu2016ssd} to estimate the 6D object pose, turning the regression into a classification problem. 
In their follow-up work~\cite{manhardt2019explaining}, Manhardt \etal leverage multiple hypotheses to improve robustness towards ambiguities.
In~\cite{manhardt2018deep}, the authors leverage ideas from projective contour alignment to estimate the pose.
A few other works also make use of the point-matching loss to directly optimize for pose in 3D~\cite{xiang2017posecnn,li2019deepim,labbe2020cosypose}. Finally, \cite{hu2020single} and \cite{GDRN} both establish 2D-3D correspondences but attempt to learn the P$n$P paradigm in an end-to-end fashion.

The next branch employs latent embedding for pose estimation. 
These learned embeddings can be then leveraged for retrieval during inference.
Specifically, inspired by~\cite{Wohlhart2015,kehl2016deep}, Sundermeyer \etal~\cite{sundermeyer2018implicit} utilize an Augmented AutoEncoder (AAE) to learn a low-dimensional pose embedding. 
After localizing the object in image space using a 2D object detector~\cite{liu2016ssd,lin2017focal}, the latent representation of the detection is calculate and compared against a pre-computed codebook to retrieve the pose. 
To further improve scalability to multiple objects, the authors of~\cite{sundermeyer2020multi} propose to employ a single-shared encoder together with separate decoders for each object.

Finally, the last branch is grounded on establishing 2D-3D correspondences, before solving for the pose using RANSAC/P$n$P. 
Thereby, some works propose to regress the 2D projections of the 3D bounding box corners~\cite{rad2017bb8,tekin18_yolo6d}. 
To increase the robustness of these correspondences, Hu \etal predicts multiple hypotheses on the basis of segmented super-pixels. 
Nevertheless, most recent methods utilize 2D-3D correspondences with respect to the 3D model rather than the 3D bounding box. 
Peng \etal~\cite{peng2019pvnet} demonstrate that keypoints away from the object surface induce larger errors and, therefore, instead sample several keypoints on the object model based on farthest point sampling. 
HybridPose~\cite{song2020hybridpose} follows and develops \cite{peng2019pvnet} by introducing hybrid representations.
Noteworthy, the majority of works within this branch, however, establishes dense 2D-3D correspondences~\cite{zakharov2019dpod,li2019cdpn,park2019pix2pose,hodan2020epos}. 
They are among the best-performing methods on several challenging benchmarks.
\begin{figure*}[t]
\centering
\includegraphics[width=0.98\textwidth]{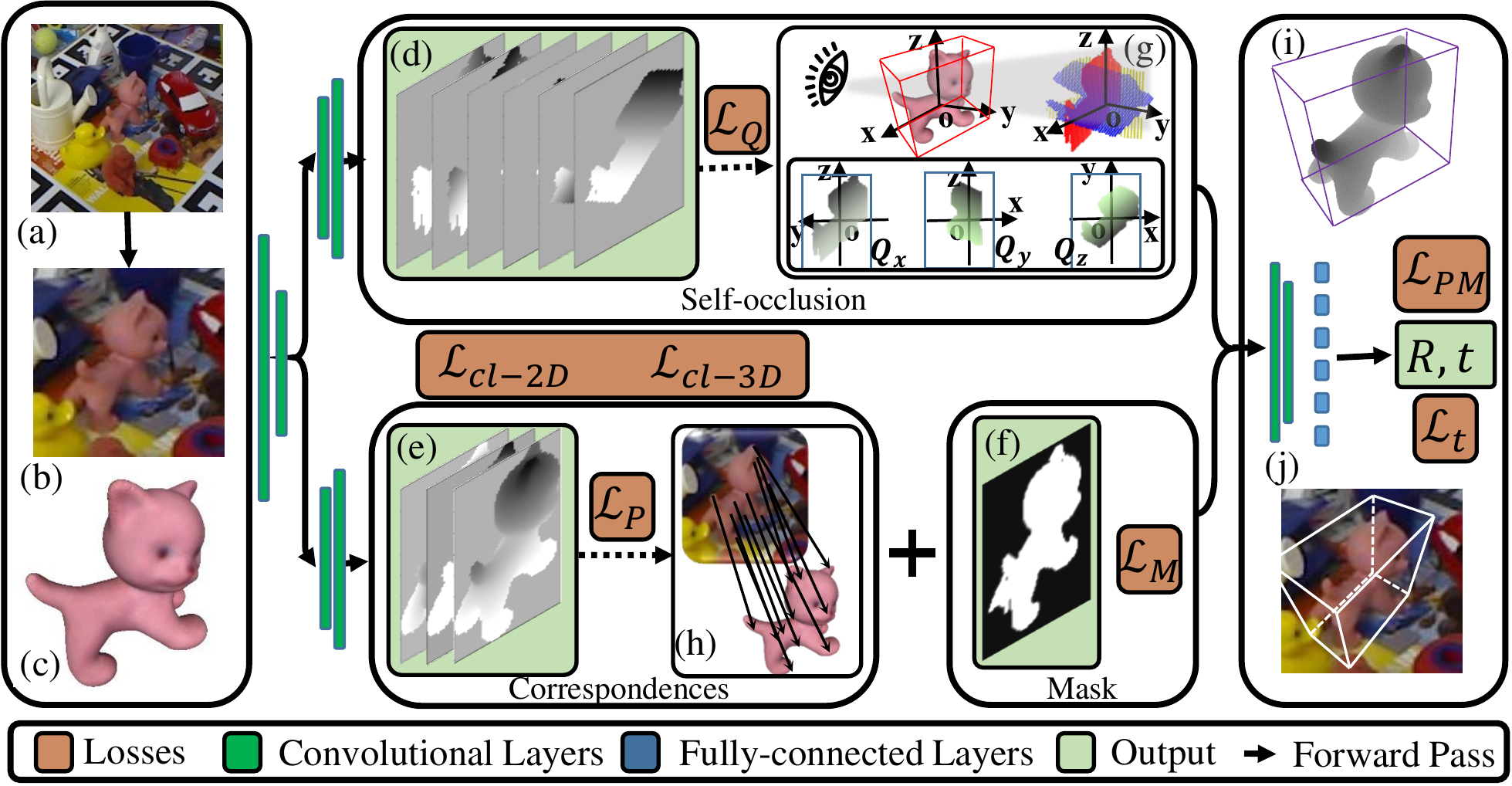} 
\caption{\textbf{Schematic overview of the proposed SO-Pose framework.}
Given an input image (a) and 3D model (c), we first use an off-the-shelf object detector~\cite{redmon2016you, redmon2018yolov3} to crop the object of interest (b) from (a). 
Afterwards, (b) is fed into our encoder for high-level feature extraction. 
These features are then separately processed by two individual decoder networks to predict a two-layer representation.
Thereby, while the first branch outputs a self-occlusion map (d), the latter branch estimates 2D-3D point correspondences (e) and object mask (f).
In (h), we demonstrate the established 2D-3D correspondences between the visible surface of the object and the 3D model.
A detailed illustration of self-occlusion is shown in (g).
For a visible point on the object surface, it occludes coordinate planes $o-yz$, $o-xz$, $o-xy$ at $Q_x$, $Q_y$ and $Q_z$. 
Finally, we feed the self-occlusion map (d), together with the 2D-3D point matching field (e) to the pose estimator block that predicts the final 6D pose.
Exemplary output for depth map and the rendered bounding box using the estimated pose are shown in (i) and (j), respectively. 
$\ll_{*}$ represent loss terms used in the training process.
}
\label{fig_pipeline}
\end{figure*}

\section{Methodology}

Given an image $I$, we leverage a neural network to learn a mapping $f(\cdot)$ from $I$ to the relative 3D rotation $R$ and translation $t$, transforming the target object from the object frame to the camera frame,
\begin{equation}
    \label{goal}
R,t=f(I;\Theta ),
\end{equation}
with $\Theta$ denoting the trainable parameters of the utilized network. 
In a cluttered environment, the available object information is oftentimes severely limited due to (self-) occlusion.
Moreover, directly regressing the 3D rotation parameters under occlusion has proven to be challenging~\cite{hodan2018bop}.
Inspired by multi-layer models in 3D reconstruction~\cite{shin2019multilayer}, we propose to combine visible 2D-3D correspondences with invisible self-occlusion information to establish a two-layer representation for objects in 3D space, in an effort to capture more complete geometric features than single-layer approaches relying only on correspondences~\cite{hu2020single,GDRN}. 
Thereby, we enforce two cross-layer consistencies to align self-occlusion, correspondence field and 6D pose, 
to reduce the influence of noise and, thus, enhance pose estimation under various challenging external influences.
The overall architecture of SO-Pose is illustrated in Figure~\ref{fig_pipeline}.


\subsection{Self-occlusion for Robust Pose Estimation} 
\begin{figure}[t!]
\centering
\includegraphics[width=0.46\textwidth]{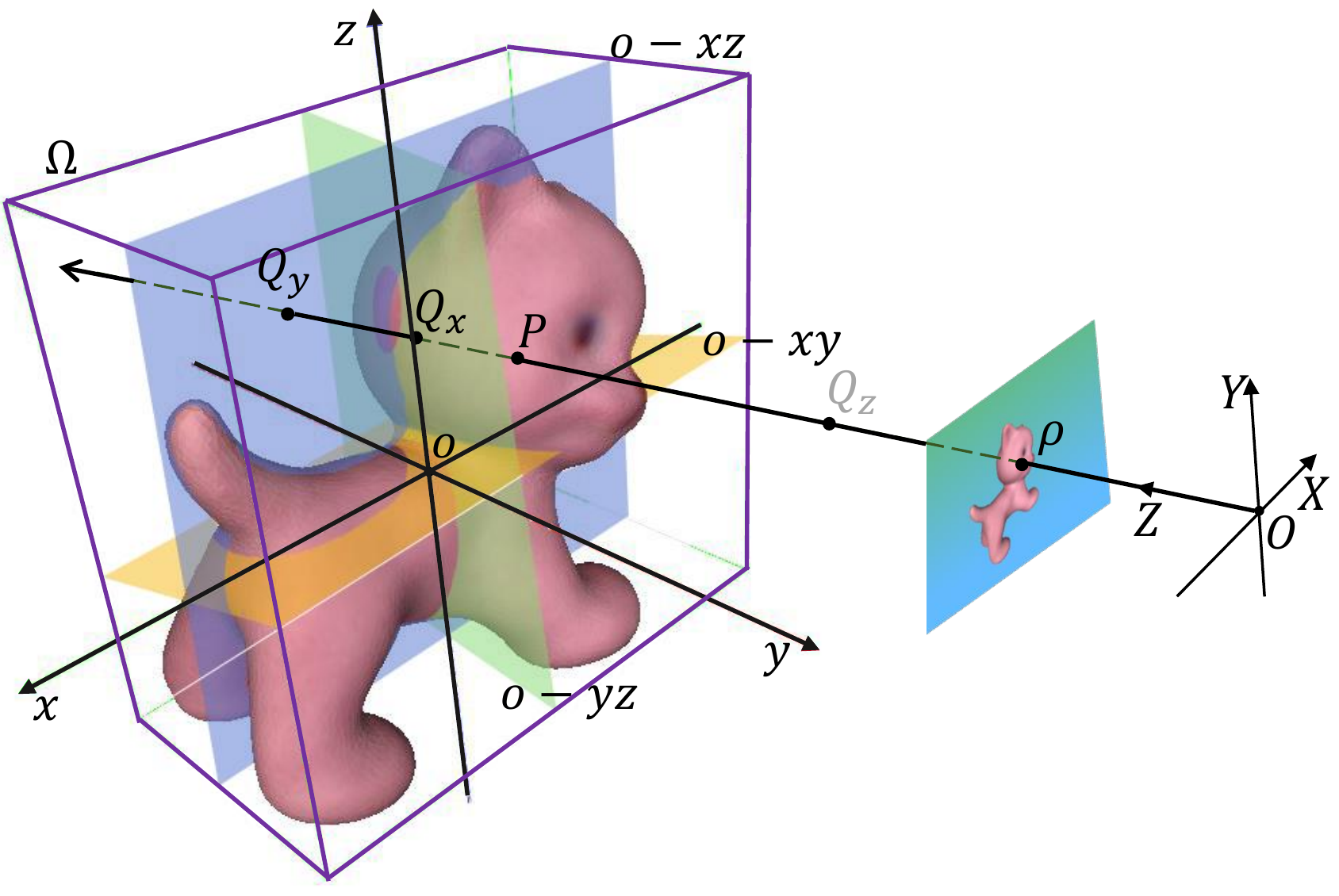} 
\caption{\textbf{Demonstration of self-occlusion.}
The ray $OP$ from camera center $O$ towards the visible point $P$ intersects the object coordinate planes $o-yz, o-xz, o-xy$ at 3 points $Q_x$, $Q_y$ and $Q_z$.
We only consider the points inside the pre-defined region $\Omega$ for training stability, thus $Q_z$ would be removed in this example.}
\label{fig_pro_selfocc}
\end{figure}

The vast majority of CNN-based 6D pose estimation approaches focus only on the visible part of an object while discard the occluded part~\cite{hu2019segmentation,hodan2020epos,GDRN}. 
Nevertheless, in complex environments, the visible region of an object is oftentimes very limited or only exhibits little amount of textured information. Hence, single-layer representations can not encode the geometric features of the object completely and accurately, inducing ambiguities for 6D pose.
Similar to multi-layer model in 3D reconstruction \cite{shin2019multilayer}, we attempt to leverage self-occlusion information to obtain a richer representation of the 3D object. 
As shown in Fig.~\ref{fig_pipeline} (d) and (e), we combine self-occlusion with estimated 2D-3D correspondences to establish a novel two-layer representation for describing the pose of an object in 3D space.
For better understanding, imagine a ray emitted from the camera center and passing through the object.
This ray intersects the object surface at multiple different points, of which the first one is visible whereas all others are self-occluded. 
In contrast to \cite{shin2019multilayer} which records the coordinates of self-occluded points, we instead note the coordinates of the intersections between each ray and the coordinate planes of the object. 
As illustrated in Fig.~\ref{fig_pro_selfocc}, the ray $OP$ intersects the object coordinate frames $o-yz, o-xz, o-xy$ at points $Q_x, Q_y, Q_z$.
For an object $obj$, we combine $P, Q=\{Q_x, Q_y, Q_z\}$ to represent its two-layer model,
\begin{equation}
    \label{twolayermodel}
    \begin{matrix}
obj \coloneqq \{P,Q\} & 
P \in V,
\end{matrix}
\end{equation}
with $V$ denoting the visible points of the current view \wrt the camera coordinate system.
Notice that $Q$ can be derived analytically from $P$, knowing the rotation $R$ and translation $t$. 
Projecting $P$ onto the 2D image plane, we obtain
\begin{equation}
    \label{camera_pro}
    \begin{matrix}
\rho =\frac{1}{Z_P}KP,
\end{matrix}
\end{equation} with $K$ describing the camera intrinsic matrix and $P=\left [ X_P, Y_P, Z_P \right ]^T$ denoting the visible 3D point. 
Further, the object coordinate plane \wrt the camera coordinate system is defined as
\begin{equation}
    \label{planeeq}
    (Rn_{*})^TX = (Rn_{*})^Tt.
\end{equation}
where $X$ represents a 3D point on the corresponding coordinate plane given
\begin{equation}
    \label{n}
    n_{*}=\left\{\begin{matrix}
n_x=\left [ 1, 0, 0 \right ]^T & X \in o-yz\\ 
n_y=\left [ 0, 1, 0 \right ]^T & X \in o-xz\\ 
n_z=\left [ 0, 0, 1 \right ]^T & X \in o-xy
\end{matrix}\right.
\end{equation}

From this we can derive $Q_x$, lying on the plane $o-yz$ intersected by the ray $OP$, as follows,
\begin{equation}
    \label{Q}
    Q_x = \frac{(Rn_x)^Tt}{(Rn_x)^T(K^{-1}\rho)}K^{-1}\rho
\end{equation}
Substituting $n_x$ with $n_y$ or $n_z$ in Eq.~\ref{Q}, we can respectively derive $Q_y$ or $Q_z$.

Since $P$ and $Q$ are represented \wrt the camera coordinate system, their corresponding coordinates \wrt the object coordinate system are calculated as $P_0 = R^TP-R^Tt$ and $Q_0 = R^TQ-R^Tt$.
Notice that we normalize $P_0$ and $Q_0$ based on the object diameter to stabilize optimization. 
Notably, as the ray passing through the camera center $O$ and a visible point $P$ may be parallel to one of the object coordinate planes, it can occur that the ray never intersects this plane. 
Therefore, to circumvent these cases and increase robustness, we only consider intersections inside the minimum bounding cuboid $\Omega$ of the object, as visualized in Fig.~\ref{fig_pro_selfocc}.

According to the definition of self-occlusion in Eq.~\ref{Q}, our two-layer representation exhibits 3 advantages over single-layer approaches.
First, the self-occlusion coordinate $Q$ is analytically derived with rotation and translation parameters for each visible point, independent of the object surface.
Thus it eliminates errors arising from rendering.
Moreover, since the self-occlusion coordinate $Q_0$ lies on a coordinate plane and has therefore only 2 degrees-of-freedom, we also only require to predict 2 values to represent $Q_0$. Hence, this acts as a regularization term which can reduce the influence of noise.
Finally, since $P$ and $Q$ lie on the same line, we can derive several cross-layer consistencies which align self-occlusion, 2D-3D correspondences and 6D pose, increasing significantly accuracy and robustness of SO-Pose, especially in challenging environments.

\subsection{Cross-layer Consistency}
Leveraging the estimated two-layer representation for objects in 3D space, we enforce two cross-layer consistency loss terms to jointly align self-occlusion, correspondence field and 6D pose parameters.
When substituting Eq.~\ref{camera_pro} into Eq.~\ref{Q} and rearranging, we obtain that
\begin{equation}
    \label{rtprpq}
   (Rn)^TtP=(Rn)^TPQ.
\end{equation}
From this we enforce our first cross-layer consistency as
\begin{equation}
    \label{ct_loss_3D}
    \begin{split}
    \ll_{cl-3D}&=\frac{1}{|Q_0|} \sum_{P\in V, Q_0 \in \Omega}\|(Rn)^Tt(RP_0+t) \\
    &-(Rn)^T(RP_0+t)(RQ_0+t) \|_1,
    \end{split}
\end{equation}
with $\|*\|_1$ representing the $L1$ loss and
$|Q_0|$ denoting the number of intersections within $\Omega$. 
Eq.~\ref{ct_loss_3D} jointly aligns and refines 2D-3D correspondences $P$, self-occlusion $Q$ and pose $R, t$ in 3D space, based on the definition of $Q$ in Eq.~\ref{Q}.
 
While the first cross-layer consistency is enforced in 3D space, we employ the second cross-layer loss on the 2D image plane.
Since $P$ and $Q$ lie on the same ray, their projections describe the same point $\rho$ on the image plane. 
As a consequence we can derive our 2D consistency term as follows,
\begin{equation}
    \label{ct_loss_2D}
    \ll_{cl-2D}=\frac{1}{|Q_0|} \sum_{P\in V, Q_0 \in \Omega}(\|e_{PQ}\|_1+\|e_{Q\rho}\|_1)
\end{equation}
with
\begin{equation}
\label{ct_loss_2D_epq}
e_{{\tiny PQ}}=\frac{1}{Z_Q}K(RQ_0+t)-\frac{1}{Z_P}K(RP_0+t)
\end{equation}
and
\begin{equation}
\label{ct_loss_2D_eqrho}
e_{{\tiny Q\rho}}=\frac{1}{Z_Q}K(RQ_0+t)-\rho.
\end{equation}
where $e_{PQ}$ forces $P$ and $Q$ to project onto the same 2D point, while $e_{Q\rho}$ forces $Q$ to project to $\rho$, the corresponding ground truth projection of $Q$.

\subsection{Overall Objective}
SO-Pose takes a single RGB image as input and directly predicts 6D pose parameters of an object in 3D space.
To establish a two-layer representation, our framework generates a correspondence field and three self-occlusion maps. %
In the following, both intermediate geometric features are concatenated and fed to the pose predictor to obtain the output pose in a fully differentiable fashion.
Our overall objective function is composed of basic terms for pose, cross-layer consistency terms, and self-occlusion term,
\begin{equation}
    \label{loss_all}
    \ll = \ll_{pose} + \ll_{cl} + \ll_{occ},
\end{equation}
with 
\begin{equation}
    \label{loss_cl}
    \ll_{cl} = \lambda_1 \ll_{cl-2D}  + \lambda_2 \ll_{cl-3D}
\end{equation}
and
\begin{equation}
    \label{loss_occ}
    \ll_{occ} = \lambda_3 \ll_{Q}.
\end{equation}

In particular, $\ll_{pose}$ is a combined loss term for correspondence field, translation parameters, visible mask, region classification and point matching as in \cite{GDRN}. We kindly refer the reader to \cite{GDRN} for more details on the pose terms.

As for self-occlusion, $\ll_Q$ is composed of two parts,
\begin{equation}
    \label{lossQ}
    \ll_Q=\ll_{Q_1} + \ll_{Q_2}.
\end{equation}
Thereby, we straightforwardly employ the $L_1$ loss according to
\begin{equation}
    \label{lossQ1}
    \ll_{Q_1}=\frac{1}{|Q_0|}\sum _{Q_0\in\Omega }\|Q_0-\hat{Q_0}\|_1,
\end{equation}
 with $\hat{Q_0}$ denoting the ground truth self-occlusion coordinates.

Similarly, also for $\ll_{Q_2}$ we directly employ the $L_1$ loss to ensure consistency after projection using  
\begin{equation}
    \label{lossQ2}
    \ll_{Q_2}= \frac{1}{|Q_0|}\sum _{Q_0\in\Omega}\|\frac{1}{Z_Q}K(\hat{R}Q_0+\hat{t})-\rho\|_1.
\end{equation}
Thereby, $\hat{R}$ and $\hat{t}$ represent the ground-truth rotation and translation.
Eq.~\ref{lossQ2} enforces that all predicted self-occlusion coordinates $Q_x$, $Q_y$ and $Q_z$ reside on the same ray with respect to $P$.

\section{Evaluation}
In this section we compare SO-Pose to current state-of-the art methods in 6D pose estimation.
We conduct extensive experiments on three challenging datasets to demonstrate the effectiveness and superiority of our approach. 
We also perform various ablation studies to verify that our two-layer model consistently surpasses the single-layer competitors.

\subsection{Network Structure}
We feed SO-Pose with a zoomed-in RGB image~\cite{li2019cdpn,GDRN} of size $256\times$256 as input and directly output 6D pose.
Similar to GDR-Net~\cite{GDRN}, we parameterize the 3D rotation using its allocentric 6D representation $R_{6d}$~\cite{kundu20183d,GDRN}, and the 3D translation as the projected 3D centroid and the object's distance\cite{li2019cdpn}.
As backbone we leverage ResNet34 \cite{he2016deep} for all experiments on the LM dataset~\cite{hinterstoisser2012}, while we employ ResNeSt50~\cite{zhang2020resnest} for the more challenging datasets, \ie~LMO~\cite{brachmann2016uncertainty} and YCB-V~\cite{xiang2017posecnn}.

After feature extraction using the aforementioned backbone, we append two decoders for estimation of self-occlusion and 2D-3D correspondences.
The first branch essentially outputs 6-channel self-occlusion maps with a resolution of $64\times$64.
The second branch predicts three different groups of intermediate geometric feature maps of size $64\times$64.
While the first group describes the visible object mask, the other two groups describe the 2D-3D correspondence field~\cite{Grabner2018} and surface region attention map as defined in \cite{GDRN}.
Finally, self-occlusion and point matching feature maps are fed into the pose regression network to predict 6D pose directly.
We adopt the identical pose regression network as in \cite{GDRN}.

\subsection{Training Details}

\paragraph{Implementation Details.} 
Our network is trained end-to-end using Ranger optimizer~\cite{ranger1,ranger2,ranger3} on a single TITAN X GPU.
We use a batch size of 24 and a base learning rate of 1e-4.
We anneal the learning rate with a cosine schedule at $72\%$ of the training phase.
Unless specified otherwise, we set $\{\lambda _1, \lambda _2,\lambda _3\}=\{1/f, 10, 1\}$, where $f$ denotes the focal length.
Moreover, to increase stability, we first train the network without $\ll_{cl-3D}$ and $\ll_{cl-2D}$ and add them after $20\%$ of the total training epochs.
During training, color augmentation and mask erosion are randomly applied to avoid overfitting similar to~\cite{sundermeyer2018implicit}.
For 2D localization we utilize Faster-RCNN~\cite{ren2015faster} on LMO and FCOS~\cite{tian2019fcos} on YCB-V.
Notice that we do not take special care of symmetric objects~\cite{GDRN, hodan2020epos} or post-refinement~\cite{li2019deepim, labbe2020cosypose}. 

\paragraph{Datasets.}~We test SO-Pose on three commonly-used datasets, \ie LM \cite{LM}, LMO \cite{Brachmann2014Learning6O} and YCB-V \cite{xiang2017posecnn}. 
LM consists of individual sequences for 13 objects undergoing mild occlusion. 
We follow \cite{GDRN,li2019cdpn} and employ $\approx$15\% of the RGB images for training and the remaining part for testing.
We additionally render 1K synthetic images for each object during training.
LMO extends LM by annotating one sequence with other 8 visible objects, often imposing severe occlusion on the objects.
Similarly, we render additional 10k synthetic images for each object.
Finally, YCB-V is a very challenging dataset exhibiting strong occlusion, clutter and several symmetric objects. 
We adopt the provided real images of 21 objects and publicly available physically-based rendered (\textit{pbr}) data for training and testing as in \cite{GDRN}.
We additionally evaluate our method following the BOP setup on LMO and YCB-V~\cite{hodan2018bop}.

\paragraph{Evaluation Metrics.}
We employ the most commonly used metrics for comparison with other state-of-the-art methods.
Thereby, \textit{ADD(-S)} \cite{LM, hodan2016evaluation} measures the percentage of transformed model points whose deviation from ground truth lies below $10\%$ of the object's diameter (0.1d).
For symmetric objects, \textit{ADD(-S)} measures the deviation to the closet model point \cite{LM, hodan2016evaluation}.
Further, $n^{\circ}n \, cm$ \cite{Shotton2013cvpr} measures the percentage of predicted 6D poses whose rotation error is less than $n^{\circ}$ and translation error is below $n \,cm$. 
On YCB-V dataset, we also compute the AUC (area under curve) of ADD-S and ADD(-S) similar to \cite{xiang2017posecnn}.
For BOP settings on LMO and YCB-V, we additionally compute $AR_{\tiny VSD}$, $AR_{\tiny MSSD}$, $AR_{\tiny MSPD}$ as proposed by~\cite{hodan2018bop}.
We also provide the average $AR$ score to compare the performance on various datasets.

\subsection{Comparison with State of the Art}
This section compares SO-Pose with other state-of-the-art methods on different datasets.

\paragraph{Results on LM.} As shown in Tab.~\ref{tab-lm}, our method consistently outperforms all baseline methods for each metric, especially in terms of $ADD(-S) \, 0.02d$ and $2^{\circ}2 \, cm$.
Compared to GDR-Net~\cite{GDRN}, under $ADD(-S)\,0.02d$, we improves from 35.3 to 45.9, up to 30\%.
Since these two strict metrics are usually utilized to measure performance in robotic grasping or other high-level tasks, the significant improvement of SO-Pose in Tab.~\ref{tab-lm} demonstrates that our method has great potential in robotic applications.

\paragraph{Results on LMO.}
We compare our method with state-of-the-art competitors in terms of $ADD(-S)$ in Tab.~\ref{new}.
When trained with \textit{real+syn}, our method achieves even comparable results to refinement-based method such as DeepIM~\cite{li2019deepim} and outperforms all other competitors.
Further, using \textit{real+pbr} data for training, our method achieves state-of-the-art performance on 5 out of 8 objects.
Thereby, our average score surpasses all other methods by a large margin, with 62.3 against $24.9-56.1$. 

\paragraph{Results on YCB-V.~}
As for YCB-V, we show our results in Tab.~\ref{Tab-ycbv}.
Using ResNeSt50~\cite{zhang2020resnest}, we outperform again all other methods under ADD(-S) and AUC of ADD-S, with 56.8 and 90.9 against second best results 53.9 and 89.8.
Under AUC of ADD(-S), we are only a little inferior to CosyPose~\cite{labbe2020cosypose}, with 83.9 compared to 84.5.
Nevertheless, whilst achieving comparable results as CosyPose, out method runs significantly faster as CosyPose is a refinement-driven method, while we only need a single forward pass to obtain the final 6D pose.

\paragraph{Results under BOP metrics.}
In Tab.~\ref{tab-bop}, we report our results under the BOP setup.
To ensure a fair comparison with the related works, on LMO we only utilize the provided pbr data for training, whereas on YCB-V, both real and pbr data are utilized~\cite{hodan2020epos,li2019cdpn,GDRN}.
For all non-refinement methods, SO-Pose again achieves superior results reporting a mean $AR$ of 0.664 compared to CDPN-v2 \cite{li2019cdpn} with 0.578 and EPOS \cite{hodan2020epos} with 0.621. 
Nevertheless, we are a little inferior to CosyPose, the overall best performing method that adopts sophisticated refinement.


\begin{table}[t!]
\centering
\scalebox{0.95}{
\begin{tabular}{c|c|c|c|c|c}
\shline
\multirow{2}{*}{Method} & \multicolumn{3}{c|}{ADD(-S)} & \multirow{2}{*}{$2^{\circ}2cm$} & \multirow{2}{*}{$5^{\circ}5cm$}\\
\cline{2-4}
& 0.02d & 0.05d & 0.1d & & \\
\hline
CDPN \cite{li2019cdpn} & - & - & 89.9 & - & 94.3\\
GDR-Net \cite{GDRN} & 35.3 & 76.3 & 93.7 & 62.1 & 95.6 \\
\hline
Ours(S) & 36.6 & 76.8 & 94.0 & 59.1 & 97.0 \\
\hline
Ours & \multirow{2}{*}{41.6} & \multirow{2}{*}{81.7} & \multirow{2}{*}{95.7} & \multirow{2}{*}{67.4} & \multirow{2}{*}{97.1} \\
(\textit{w/o} $\ll_{cl-3D}$) &  &  &  &  &  \\
\hline
Ours & \multirow{2}{*}{44.7} & \multirow{2}{*}{81.3} & \multirow{2}{*}{95.5} & \multirow{2}{*}{73.1} & \multirow{2}{*}{98.0} \\
(\textit{w/o} $\ll_{cl-2D}$) &  &  &  &  &  \\
\hline
Ours & \textbf{45.9} & \textbf{83.1} & \textbf{96.0} & \textbf{76.9} & \textbf{98.5} \\
\shline 
\end{tabular} }
\caption{\textbf{Ablation Study on LM.} We provide results of our method with different loss terms. Ours(S) stands for removing both the $\ll_{cl-3D}$ and $\ll_{cl-2D}$ terms.}
\label{tab-lm}
\end{table}
\begin{table}[t!]
\centering
\scalebox{0.92}{
\begin{tabular}{c|c|c|c|c|c}
\shline
\multirow{2}{*}{Method} & \multirow{2}{*}{P.E.} & \multirow{2}{*}{Ref.} & ADD & AUC of & AUC of\\
& & &(-S) & ADD-S & ADD(-S) \\
\hline
PoseCNN \cite{xiang2017posecnn} & 1 &  & 21.3 & 75.9 & 61.3 \\
SegDriven \cite{hu2019segpose} & 1 &  & 39.0 & - & - \\
PVNet \cite{peng2019pvnet} & M &  & - & - & 73.4 \\
S.Stage \cite{hu2020single} & M &  & 53.9 & - & - \\
GDR-Net \cite{GDRN}& 1 &  & 49.1 & 89.1 & 80.2 \\
DeepIM \cite{li2019deepim} & 1 & \checkmark & - & 88.1 & 81.9 \\
CosyPose \cite{labbe2020cosypose} & 1 & \checkmark & - & 89.8 & \textbf{84.5} \\
\hline
Ours(34) & 1 &  & 54.6 & 89.7 & 82.3 \\
Ours(50) & 1 &  & \textbf{56.8} & \textbf{90.9} & 83.9 \\
\shline 
\end{tabular}}
\caption{\textbf{Results on YCB-V.} We report the results of our method with different backbones. Ours(34) uses ResNet34 \cite{he2016deep} while Ours(50) uses ResNeSt50 \cite{zhang2020resnest}. Ref. stands for refinement.
P.E. reflects the training strategy of pose estimator, 1 represents single model for all objects while M represents one model per object.
In general, the latter strategy benefits accuracy yet limits practical use.
}
\label{Tab-ycbv}
\end{table}
\begin{table*}[t]
\centering
\begin{tabular}{c|c|c|c|c|c|c|c|c|c|c}
\shline
\multirow{3}{*}{Method} & \multicolumn{8}{c|}{\textit{w/o} Refinement} & \multicolumn{2}{c}{\textit{w/} Refinement}\\
\cline{2-11}
& PoseCNN & PVNet  & S.Stage  & HybridPose  & \multicolumn{2}{c|}{GDR-Net } & \multicolumn{2}{c|}{Ours} & DPOD  & DeepIM  \\
& \cite{xiang2017posecnn} & \cite{peng2019pvnet} & \cite{hu2020single} & \cite{song2020hybridpose} & \multicolumn{2}{c|}{\cite{GDRN}} & \multicolumn{2}{c|}{(SO-Pose)} & \cite{zakharov2019dpod} & \cite{li2019deepim}\\
\hline
P.E. & 1 & M & M & M & 1 & 1 & 1 & 1 & 1 & 1 \\
\hline
Training & real & real & real & real & real & real & real & real & real & real\\
Data & +syn & +syn & +syn & +syn & +syn & +pbr & +syn & +pbr & +syn & +syn \\
\hline
Ape & 9.6 & 15.8 & 19.2 & 20.9 & 41.3 & 44.9 & 46.3 & 48.4 & - & \textbf{59.2} \\
Can & 45.2 & 63.3 & 65.1 & 75.3 & 71.1 & 79.7 & 81.1 & \textbf{85.8} & - & 63.5 \\
Cat & 0.9 & 16.7 & 18.9 & 24.9 & 18.2 & 30.6 & 18.7 & \textbf{32.7} & - & 26.2 \\
Driller & 41.4 & 65.7 & 69.0 & 70.2 & 54.6 & 67.8 & 71.3 & \textbf{77.4} & - & 55.6 \\
Duck & 19.6 & 25.2 & 25.3 & 27.9 & 41.7 & 40.0 & 43.9 & 48.9 & - & \textbf{52.4} \\
$Eggbox^{*}$ & 22.0 & 50.2 & 52.0 & 52.4 & 40.2 & 49.8 & 46.6 & 52.4 & - & \textbf{63.0} \\
$Glue^{*}$ & 38.5 & 49.6 & 51.4 & 53.8 & 59.5 & 73.7 & 63.3 & \textbf{78.3} & - & 71.7 \\
Holep. & 22.1 & 36.1 & 45.6 & 54.2 & 52.6 & 62.7 & 62.9 & \textbf{75.3} & - & 52.5 \\
\hline
Mean & 24.9 & 40.8 & 43.3 & 47.5 & 47.4 & 56.1 & 54.3 & \textbf{62.3} & 47.3 & 55.5 \\
\shline
\end{tabular}
\caption{\textbf{Comparison with state-of-the-art methods on LMO.} We list the Average Recall of ADD(-S). $(*)$ denotes symmetric objects.}
\label{new}
\end{table*}


\begin{table*}[t]
\centering
\begin{tabular}{c|c|c|ccc|ccc|c}
\shline
\multirow{2}{*}{Method} &\multirow{2}{*}{P.E.}&\multirow{2}{*}{Ref.} & \multicolumn{3}{c|}{\textbf{LMO}} & \multicolumn{3}{c|}{\textbf{YCB-V}} & \textbf{Mean}  \\
\cline{4-9}
& & & $AR_{VSD}$ & $AR_{MSSD}$  & $AR_{MSPD}$ & $AR_{VSD}$  & $AR_{MSSD}$ & $AR_{MSPD}$ &  \textit{\textbf{AR}} \\
\hline
CosyPose \cite{labbe2020cosypose} & 1 & \checkmark & \textbf{0.480} & \underline{0.606} & {0.812} & \textbf{0.772} & \textbf{0.842} & \textbf{0.850} & \textbf{0.727}  \\
\hline
EPOS \cite{hodan2020epos} & 1 & & 0.389 & 0.501 & 0.750 & 0.626 & 0.677 & \underline{0.783} &  0.621 \\
PVNet \cite{peng2019pvnet} & M &  & 0.428 & 0.543 & 0.754 & - & - & - & -  \\
CDPN-v2 \cite{li2019cdpn} & M &  & \underline{0.445} & \textbf{0.612} & \underline{0.815} & 0.396 & 0.570 & 0.631 & 0.578  \\
GDR-Net \cite{GDRN} & 1 &  & -& -& -& 0.584 & 0.674 & 0.726 & -  \\
Ours & 1 &  & 0.442 & 0.581 & \textbf{0.817} & \underline{0.652} & \underline{0.731} & 0.763 & \underline{0.664}  \\
\shline
\end{tabular}
\caption{\textbf{Comparison with state-of-the-art methods on LMO and YCB-V under BOP metrics.} 
We provide results for $AR_{VSD}$, $AR_{MSSD}$ and $AR_{MSPD}$ on LMO and YCB-V.
Mean $AR$ represents the overall performance on these two datasets as the average over all $AR$ scores.
Overall best results are in bold and the second best results are underlined.}
\label{tab-bop}
\end{table*}

\begin{figure*}[t]
\centering
\includegraphics[width=0.98\textwidth]{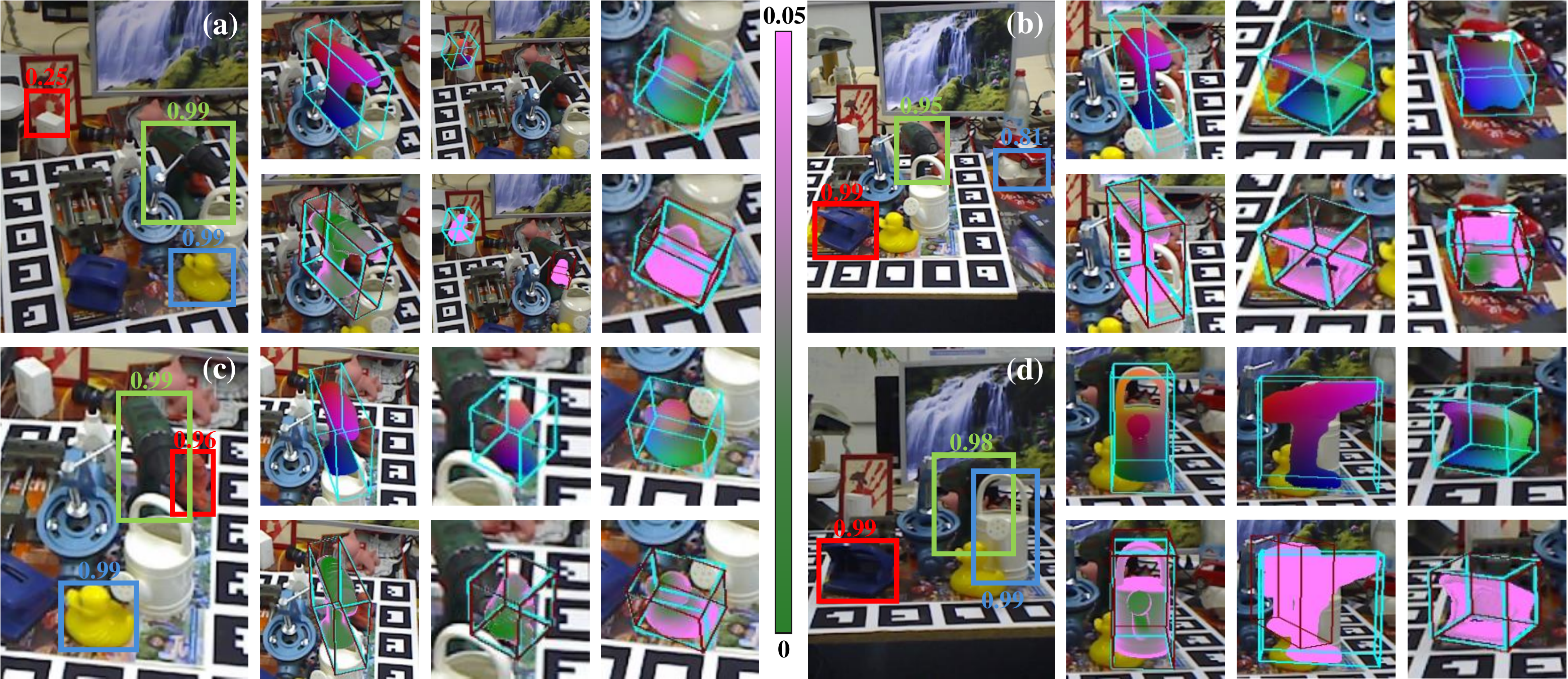} 
\caption{\textbf{Qualitative results on LMO.} We presents 4 examples of 6D pose estimation results. 
Given an input image, we first show the 2D detection results with confidence scores on top of the bounding boxes. 
Next to it we show the predicted 2D-3D matching (rendered with the predicted pose, color-coded) and object bounding box (estimated with the predicted pose, light blue) are shown on the top row.
Correspondingly, we demonstrate the error of the predicted 2D-3D matching in the second row, from green to pink as illustrated in the middle line.
We also visualize the ground truth bounding boxes (red). }
\label{fig-lmo_qua}
\end{figure*}

\begin{table}[h]
\centering
\tablestyle{8pt}{1.1}
\begin{tabular}{x{84}|ccc}
\shline
Method  & ADD(-S) & $2^{\circ}2cm$ & $5^{\circ}5cm$ \\
\hline
CDPN\textsuperscript{*} & 92.82 & 68.56 & 97.03 \\
\tabincell{c}{CDPN\textsuperscript{*}-ours (\textit{$w/o$} $\ll_{cdpn}$)} & 93.09 & 69.37 & 97.06 \\
\tabincell{c}{CDPN\textsuperscript{*}-ours (\textit{$w/$} $\ll_{cdpn}$)} & \textbf{94.77} &	\textbf{71.33} &\textbf{97.07} \\
\shline
\end{tabular}
\caption{\textbf{Evaluation of our two-layer model o top of another baseline method CDPN~\cite{li2019cdpn}.} We update the original CDPN to CDPN\textsuperscript{*}. As for SO-Pose, we integrate our self-occlusion branch into the CDPN structure. Since CDPN predicts rotation with RANSAC/P$n$P, we rearrange Eq.~\ref{ct_loss_3D} and Eq.~\ref{ct_loss_2D} to derive a new loss term $L_{cdpn}$, as defined in Eq.~\ref{losscdpn}. }
\label{tab-cdpn}
\end{table}

\begin{figure}[t!]
\centering
\includegraphics[width=0.40\textwidth]{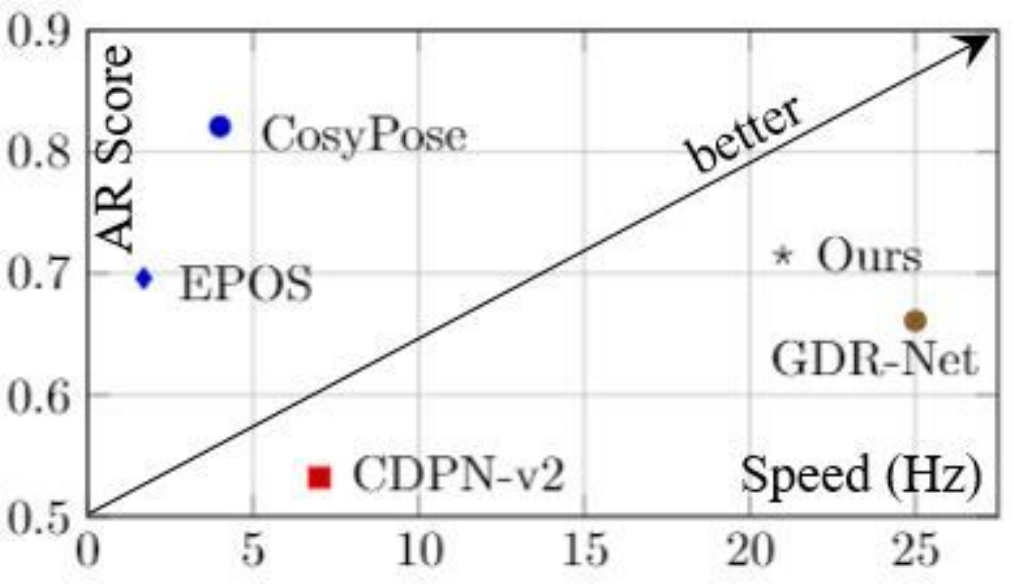} 
\caption{\textbf{Comparison of running speed (Hz) and AR score on YCB-V dataset}.
We compare our method with CosyPose~\cite{labbe2020cosypose}, GDR-Net~\cite{GDRN}, CDPN-v2~\cite{li2019cdpn} and EPOS~\cite{hodan2020epos}.
Along the direction of the arrow, method performs better, achieving higher accuracy in less inference time.
}
\label{runtime}
\end{figure}
\begin{figure}[t!]
\centering
\includegraphics[width=0.48\textwidth]{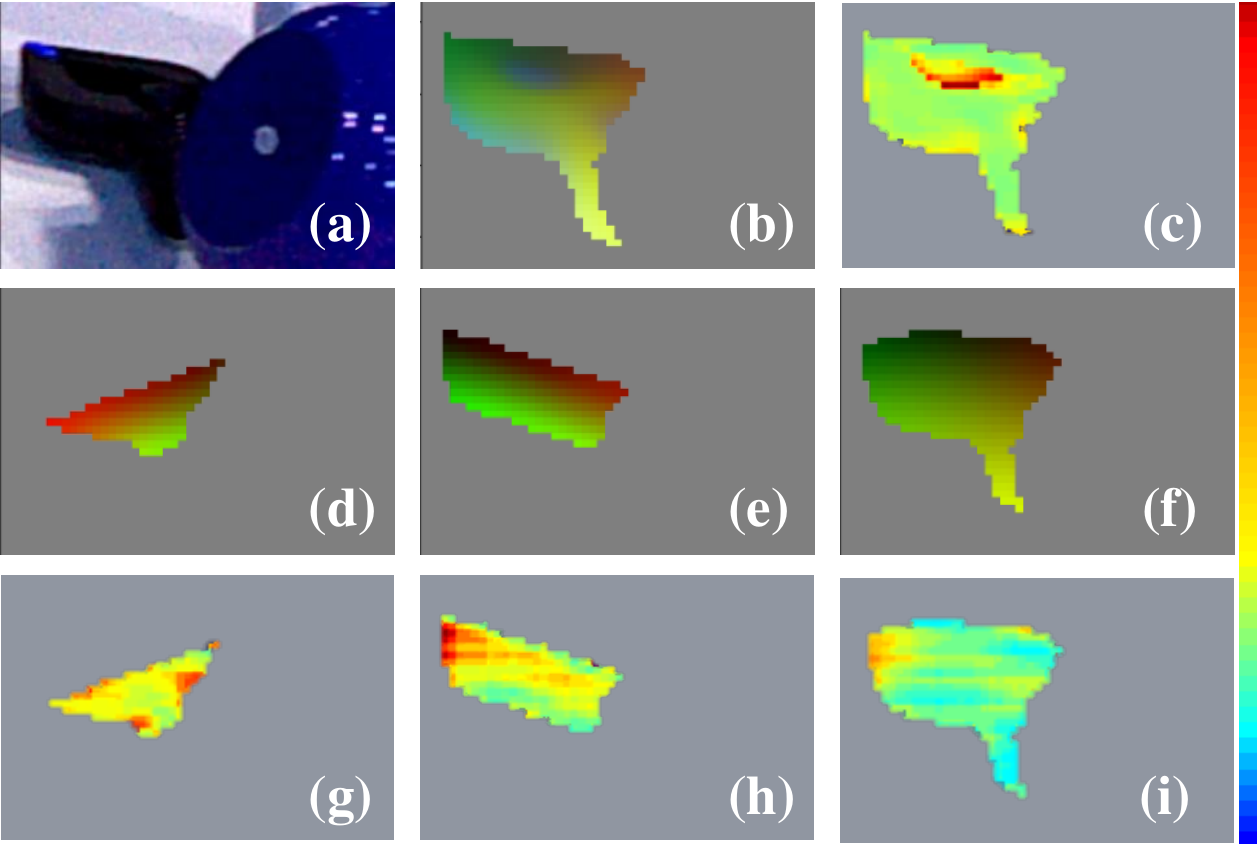} 
\caption{\textbf{Demonstration of the predicted two-layer model.} 
For object (a), we demonstrate its 2D-3D point matching in (b) and self-occlusion coordinates in (d), (e) and (f).
(c), (g), (h), (i) are corresponding error maps of (b), (d), (e), (f). 
The color bar on the right side indicates the color-coded error for error maps.
We normalize the error maps for better visualization, thus from the bottom to top of the color bar, the error ranges from 0 to 1.
}
\label{fig-direct-P}
\end{figure}

\subsection{Ablation Study}
We conduct several ablations on each dataset. 
Thereby, except for ablated terms, we leave all other terms unchanged using the values from the experimental setup.

\paragraph{Effectiveness of cross-layer consistency.}
In Tab.~\ref{tab-lm} we demonstrate the effectiveness of the proposed cross-layer consistency terms on LM. 
We gradually remove $L_{cl-3D}$ and $L_{cl-2D}$ to observe their impact on the 6D pose w.r.t ADD(-S), $2^{\circ}2\,cm$ and $5^{\circ}5\, cm$. 
Thereby, we can observe that the accuracy decreases when removing either loss term, verifying the usefulness of our cross-layer consistencies.

\paragraph{Benefits of employing self-occlusion.}
To show that self-occlusion consistently improves pose quality, we additionally adopted the two-stage method CDPN~\cite{li2019cdpn} to also incorporate our two-layer representation.
As CDPN is grounded on RANSAC/P$n$P to extract the 3D rotation from 2D-3D correspondences, we slightly adjust our cross-layer consistency terms as follows,
\begin{equation}
    \label{losscdpn}
    \ll_{cdpn}=\ll_{cl-3D}(R\rightarrow \hat{R}) +\ll_{cl-2D}(R\rightarrow \hat{R})
\end{equation}

Essentially, the consistency term $\ll_{cdpn}$ is computed by replacing the predicted rotation $R$ in $\ll_{cl-3D}$ and $\ll_{cl-2D}$ with the actual ground truth rotation $\hat{R}$.
Except for $\ll_{cdpn}$, all original loss terms in CDPN are preserved.
As shown in Tab.~\ref{tab-cdpn}, after introducing our two-layer model into CDPN, the performance improves again significantly for all metrics.
This clearly demonstrates the generalizability of our proposed two-layer model.

\paragraph{Impact of different backbones.}
We report the results of our method with ResNet34 \cite{he2016deep} and ResNeSt50 \cite{zhang2020resnest} as backbone in Tab.~\ref{Tab-ycbv}.
Although the performance degrades slightly after changing ResNeSt50 to ResNet34, our method still outperforms most state-of-the-art methods, proving its efficacy regardless of the employed backbones.

\subsection{Runtime Analysis}
Given a 640$\times$480 image from YCB-V with multiple objects, our method takes about 30ms to handle a single object and 50ms to process all objects in the image on an Intel 3.30GHz CPU and a TITAN X (12G) GPU. 
This includes the additional 15ms for 2D localization using Yolov3~\cite{redmon2018yolov3}.
As shown in Fig.~\ref{runtime}, we demonstrate the Speed-AR score figure on YCB-V.
Our method achieves second best results (AR: 0.715) in real time, which further verifies the great potential of our method in practical use.

\subsection{Qualitative evaluation}
We provide qualitative results for LMO in Fig.~\ref{fig-lmo_qua} and YCB-V in Fig.~\ref{fig-direct-P}.
In particular, in Fig.~\ref{fig-lmo_qua}, we show four exemplary results for pose estimation together with the corresponding error maps for rendered 2D-3D correspondences with the estimated pose.
In Fig.~\ref{fig-lmo_qua} (a) ape, due to wrong detection, the predicted 6D pose diverges from the ground truth completely.
In (d) driller, we demonstrate wrong prediction of 6D pose due to strong occlusion.
Finally in Fig.~\ref{fig-direct-P}, we show the normalized two-layer representation of an object from YCB-V as predicted by our model.
For more qualitative results, please refer to the Supplementary Material.

\section{Conclusion}
In this paper, we present a novel two-layer model that combines 2D-3D point correspondences and self-occlusion information to encapsulate explicitly the spatial cues of 3D object. 
Then based on the two-layer model, we establish SO-Pose, an end-to-end 6D pose regression framework that achieves significant improvements on various challenging datasets over other single-layer competitors.
The experimental evaluation also demonstrates that our two-layer model is applicable to a wide range of 6D pose estimation frameworks and can consistently benefit the performance.

In the future, we plan to focus on integrating the two-layer model into self-supervised 6D pose estimation and category-level unseen object analysis.

{\small
\bibliographystyle{ieee_fullname}
\bibliography{egbib}
}

\end{document}